%% file: main.tex
\documentclass{article}

\usepackage[final]{corl_2020} 
\usepackage{hyperref}
\usepackage{amssymb}
\usepackage{amsmath,amsfonts}
\usepackage{algorithmic}
\usepackage{gensymb}
\usepackage{array}
\usepackage{pifont}
\usepackage{hhline}
\usepackage{chngpage}
\usepackage{adjustbox}
\usepackage{caption}
\usepackage{wrapfig,lipsum,booktabs}
\usepackage{makecell}
\usepackage{multirow}
\usepackage{graphicx} 
\usepackage{authblk}

\graphicspath{ {images/} }

\def\oursfull{Grasp Detection Network}
\def\ours{GDN}

\makeatletter
\renewcommand\AB@affilsepx{, \protect\Affilfont}
\makeatother

\title{GDN: A Coarse-To-Fine (C2F) Representation for End-To-End 6-DoF Grasp Detection}

\author[ ]{Kuang-Yu Jeng}
\author[ ]{Yueh-Cheng Liu}
\author[ ]{Zhe Yu Liu}
\author[ ]{Jen-Wei Wang}
\author[ ]{Ya-Liang Chang}
\author[ ]{Hung-Ting Su}
\author[ ]{Winston H. Hsu}
\affil[ ]{National Taiwan University}

\begin{document}
\maketitle


\begin{abstract}

    We proposed an end-to-end grasp detection network, \oursfull{} (\ours{}), cooperated with a novel coarse-to-fine (C2F) grasp representation design to detect diverse and accurate 6-DoF grasps based on point clouds. Compared to previous two-stage approaches, which sample and evaluate multiple grasp candidates, our architecture is at least 20 times faster. It is also 8\% and 40\% more accurate in terms of the success rate in single object scenes and the complete rate in clutter scenes, respectively. Our method shows superior results among settings with different numbers of views and input points. Moreover, we propose a new AP-based metric which considers both rotation and transition errors, making it a more comprehensive evaluation tool for grasp detection models.
    
\end{abstract}



\section{Introduction}
	
	Grasp detection is an important problem for robotic grasping. Nowadays, robots are required to work in unstructured environments, such as warehousing or households. Therefore, it is necessary for robots to have dexterity in order to handle these difficult situations. Researchers \citep{lenz2015deep, mahler2017dex, redmon2015real} propose algorithms that can detect top-down grasps. Nonetheless, the 3-DoF nature of the top-down grasps does not allow robots to grasp target objects from different orientations, which is necessary to prevent collision with adjacent objects in many situations.
	
\begin{figure}[h]
    \centering
    \includegraphics[width=0.9\textwidth]{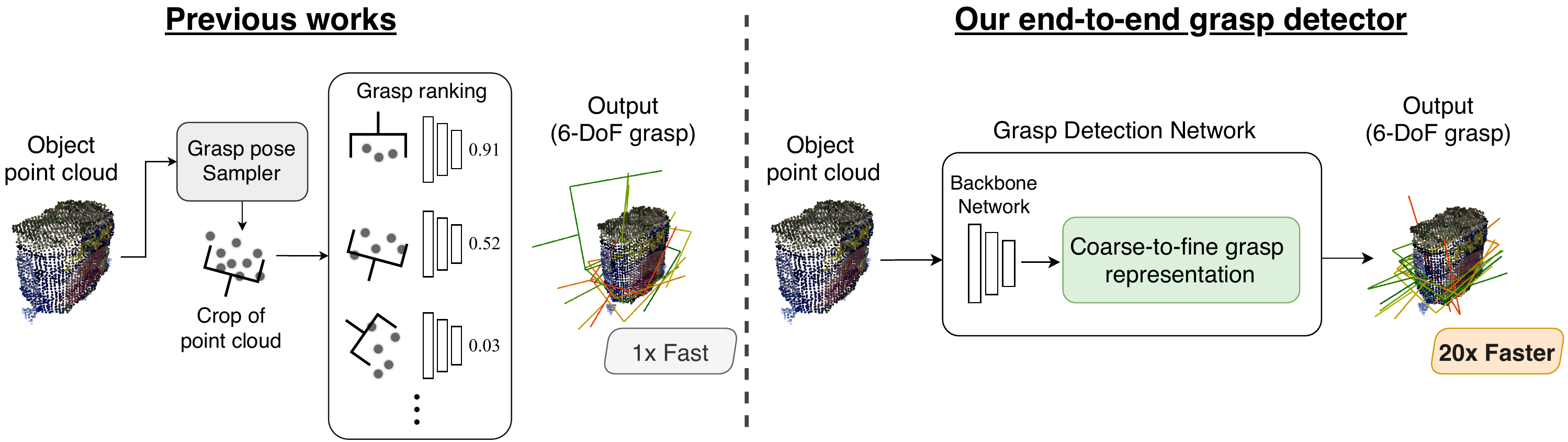}
    \caption{Previous works will sample multiple grasp candidates and evaluate each of them independently. Our end-to-end 6-DoF grasp detector directly generates multiple grasp configurations in one stage, which is faster and more accurate. In addition, we propose a coarse-to-fine grasp representation to better model the orientation of grasp poses with high diversity.}
    \label{fig:1}
    \end{figure}

    Thus, many researches have focused on detection algorithms that can determine 6-DoF grasps, which is illustrated in Figure~\ref{fig:1}. Many works \citep{ten2017grasp, liang2019pointnetgpd, qi2017pointnet, mousavian20196} solve the 6-DoF grasps with a two-stage pipeline, which samples and evaluates multiple grasp candidates separately. However, the process is time-consuming, especially when there are large number of candidates and these candidates are often essential to find a proper grasp.
    
    Recent works \citep{qin2020s4g, ni2020pointnet++} propose a single-stage network to determine the 6-DoF grasp poses and their grasp quality to mitigate the time-consuming issue. However, to avoid learning the average of ground truth grasps around one input point, these approaches need to choose only one of the ground truth grasps by heuristic for pose regression, neglecting other possible ground truth grasps with different orientations and translations. Such a strategy leads to some loss of label diversity, while the label diversity could significantly improve the accuracy for multi-trial grasps as some grasps may not be reachable due to the characteristic of the robotic arm. \citep{mousavian20196}. 
    
    
    To overcome the pose diversity and inference speed challenges mentioned above, we introduce a novel \textbf{coarse-to-fine (C2F) representation} to the single-stage end-to-end framework for 6-DoF grasp detection. Given the input point cloud, the PointNet++-based backbone will subsample them into fewer points, which we call \emph{grasp points}. Our intuition is that the desired grasp distribution will be near these subsampled points. Then, our network will output the C2F representation directly on each grasp point with shared multi-layer perceptron, while the C2F representation will be transformed into the final 6-DoF grasp pose.
    
\begin{wrapfigure}[18]{r}{0.33\textwidth}
    \vspace{-18pt}
    \includegraphics[width=0.33\textwidth]{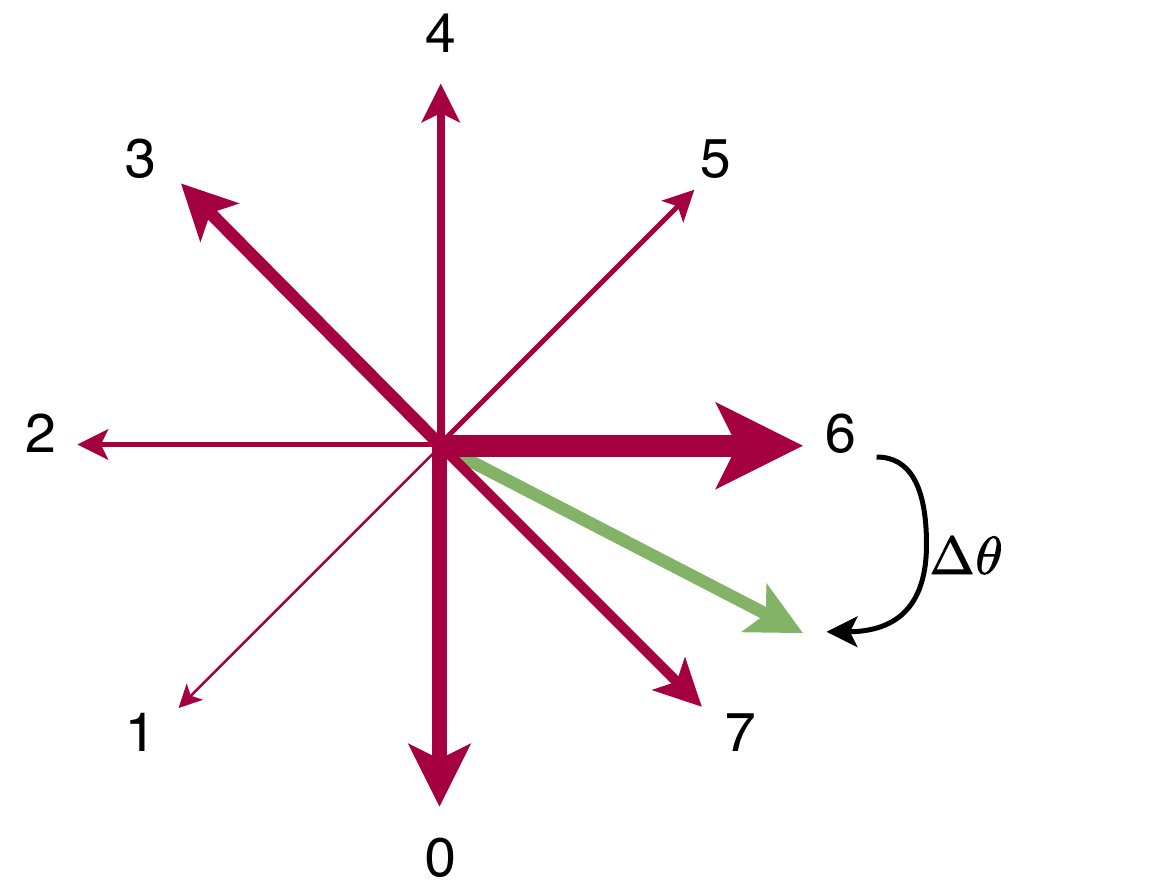}
    \caption{Illustration of coarse-to-fine process. Our approach will determine the confidence scores on each orientation (the width of each red arrow). The orientation with the highest score will be chosen (arrow 6) and the coarse pose will be fine-tuned to the exact pose (green arrow).}
    \label{fig:angle_quantization}
    \end{wrapfigure}

    The C2F representation contains the coarse part and the refinement part, which is illustrated in Figure~\ref{fig:angle_quantization}. The coarse part is a grasp confidence grid that corresponds to different grasp orientations in a quantized manner. Several coarse poses can be chosen from this grid according to their confidence scores. Thereby, our approach can detect multiple grasp poses on each grasp point and keep the pose diversity.  The refinement part is composed of residual values that are used to fine-tune each rough pose.
    
    Combining the coarse and refinement part, our C2F representation can help the network predict diverse and accurate poses in one forward pass. Due to the single-stage end-to-end network design, our approach is 20 times faster than state-of-the-art two-stage approaches, GPD \citep{ten2017grasp} and PointNetGPD \citep{liang2019pointnetgpd}, on the 38 objects from the YCB dataset \citep{calli2015ycb}. Additionally, the robot grasping experiments in simulation demonstrates that our model can achieve 95.6\% (+8\%) success rate in single object scenes and 85.33\% (+40\%) complete rate in clutter scenes.
    
    Besides, we propose a new correctness metric based on Average Precision (AP) for grasp detection models. It considers both the translation and rotation errors between the detected and ground-truth grasps, which is more comprehensive than the previous metric \citep{mousavian20196} consisting of only a translation error. In addition, the metric is independent of the robot environment and can efficiently evaluate the performance of each model because it does not depend on the experimental setup, such as gripper geometry and surface friction, during the real execution.

\section{Related Works}
\label{sec:Related Works}

\subsection{3-DoF Grasp}
\label{sec:3-DoF Grasp}

    Many researchers adopt deep learning methods for the grasp detection problem. Some of the researches propose deep learning models to detect 3-DoF grasps based on RGB or RGBD images. The proposed grasps are perpendicular to the image plane and are often depicted as rectangles on the input images. Some works \citep{lenz2015deep, mahler2017dex} have introduced deep learning models to evaluate the quality of the sampled 3-DoF grasps. However, it is time-consuming for the deep models to process the grasps. To overcome this problem, Redmon et al. \citep{redmon2015real} adopt a network that is similar to object detection models such as YOLO \citep{redmon2016you} or Faster R-CNN \citep{ren2015faster}, to determine both grasp candidates and their corresponding qualities. In addition, Zeng et al. \citep{zeng2018robotic} use semantic segmentation network to predict grasp quality on each pixel, so that the grasp distribution can be found on the image plane accordingly. Nevertheless, these image-based approaches are limited to 3-DoF grasps, but 6-DoF grasps are often required to successfully grasp some target objects. For instance, if a target is near the wall of a bin, there may not exist a 3-DoF grasp pose (top-down grasp) to pick up the object without causing a collision with the wall, while 6-DoF grasps are more possibly to do it due to the extra degrees of freedom. Therefore, many approaches tend to design deep learning models that can determine 6-DoF grasps in order to propose diverse sets of grasps.

\subsection{Two-stage 6-DoF Grasp}
\label{sec:Two-stage 6-DoF Grasp}

    One major approach for the 6-DoF grasp detection divides the problem into the generation step and the evaluation step. In the generation step, an algorithm will sample multiple grasp candidates around each object. In the evaluation step, the quality of each sampled candidate will be evaluated to determine the best grasp. Pas et al. \citep{ten2017grasp} propose an algorithm called Grasp Pose Detection (GPD) for 6-DoF grasp detection. GPD evenly samples grasp candidates over the visible object surface and adopt a Convolutional Neural Network (CNN) model to classify each grasp candidate as good or bad. They represent each grasp candidate in terms of the geometry of the object surface within the gripper and use multiple grasp images as the input to the classifier. However, the grasp images may lose some information about the 3D geometry, resulting in low classification accuracy. Therefore, Liang et al. \citep{liang2019pointnetgpd} propose PointNetGPD which utilizes the PointNet \citep{qi2017pointnet} structure to deal with the point cloud data for maintaining its 3D information.
    
    Nevertheless, there are some drawbacks in the generation step in the above two approaches. They both use a geometry-based grasp sampler for grasp generation and these kinds of grasp samplers cannot sample diverse sets of grasps around objects. Furthermore, the diversity of sampled grasps is crucial for subsequent steps because some grasp candidates might cause a collision with adjacent objects, and some might be unreachable for the robot arm. Therefore, Mousavian et al. \citep{mousavian20196} adopt Variational Auto-Encoder (VAE) as a learning-based grasp sampler for generating diverse sets of grasps. They also use another evaluation network to assess the quality of each sampled grasp. In addition, a grasp refinement step is added to the detection pipeline as the last step. They use the gradient of the evaluation network to fine-tune the pose of each sampled grasp, so the final grasp can be well aligned to target objects. Although the approach can generate diverse sets of grasps while maintaining a high success rate, the three-step pipeline is still time-consuming.

\subsection{End-to-end 6-DoF Grasp}
\label{sec:End-to-end 6-DoF Grasp}

    A few end-to-end approaches \citep{qin2020s4g, ni2020pointnet++} are proposed recently where the deep models are designed to simultaneously perform grasp pose detection and grasp quality assessment. These two tasks are often formulated as a regression problem and a classification problem respectively. However, there may exist multiple pose definitions on one grasp configuration, resulting in one-to-many mapping during the training process of the deep learning model. Because of two reasons, pose definition is not unique. The first reason is that the two-fingered parallel gripper is a symmetric object. In addition, Euler angles are not uniquely defined in some particular grasp pose. The non-uniqueness situation is often called pose ambiguity. These approaches carefully label pose values on the ground-truth grasps for avoiding this problem. However, the proposed end-to-end framework cannot detect diverse grasps because of the design of their grasp representations. Based on their representations, the networks can only determine one grasp pose around each grasp point. Nonetheless, on some grasp points, there may exist multiple ground-truth poses. These approaches may fail to find out all possible poses and may lack of diversity. Therefore, we propose a novel C2F grasp representation in an end-to-end network to maintain the pose diversity on each grasp point. 
    

\begin{figure}[h]
    \centering
    \includegraphics[width=0.85\textwidth]{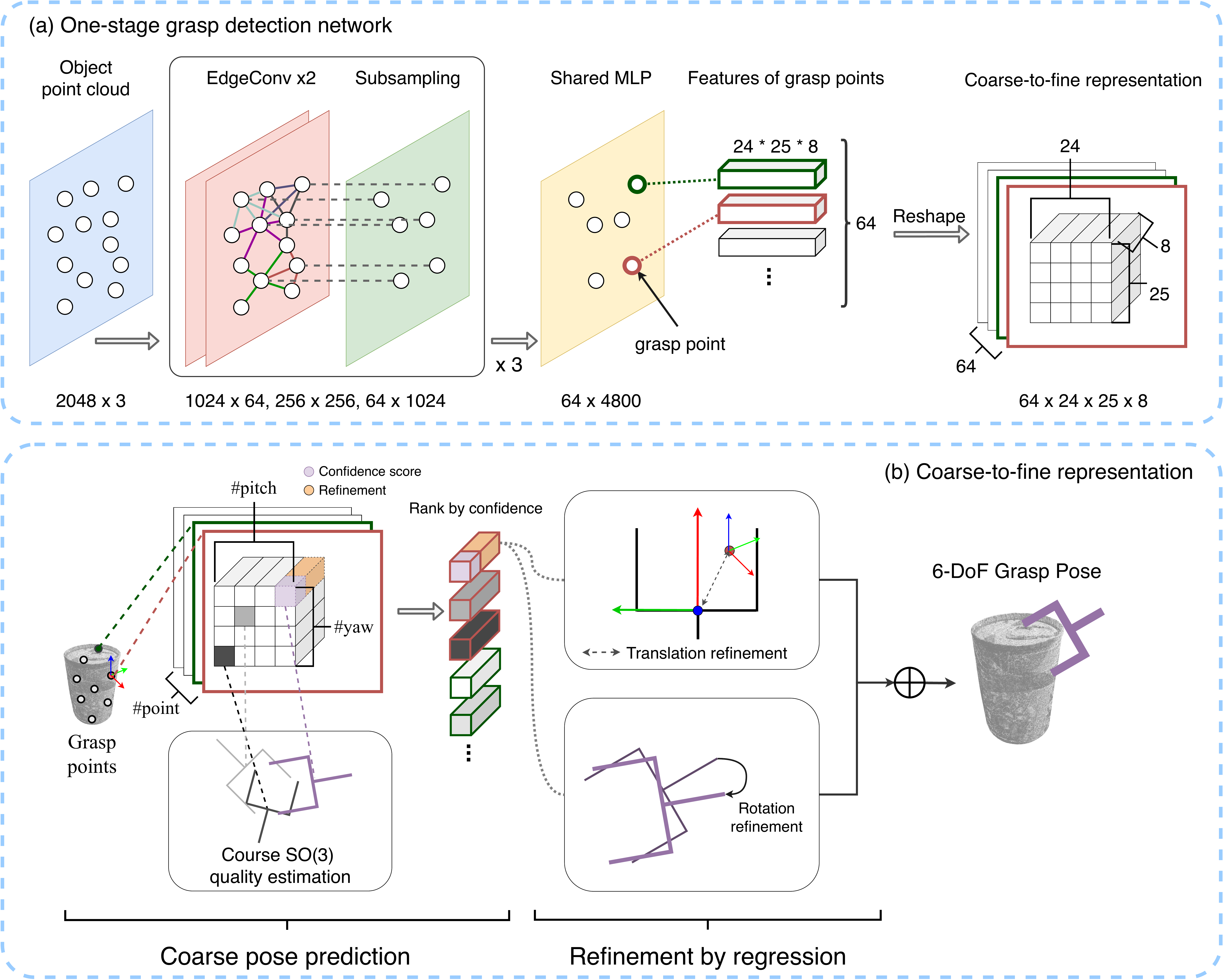}
    \caption{Our GDN predicts a 3D feature volume at every grasp points. Each feature volume represents multiple grasp configurations with respect to the position of points, and each 1D feature vectors in the feature volume represents a grasp pose with respect to the point position and the quantized coarse rotation.}
    \label{fig:2}
\end{figure}
    


\input{the_new_approach_section}
	

\section{Experiments}
\label{sec:result}

    \subsection{Evaluation Metrics}\label{sec:Evaluation metrics}
    We evaluate the grasping performance \ours{} with (1) \emph{success rate}, \emph{complete rate} which examine the reliability on grasping tasks, and (2) our proposed average precision (AP) based metrics $AP_{E}$ and $AP_H$. 
    The existing metrics such as the success and complete rates \citep{ten2017grasp, liang2019pointnetgpd, qin2020s4g, ni2020pointnet++} are sensitive to the experimental environment and need to be evaluated along with motion planning and other components. 
    Therefore, we proposed novel AP-based evaluation metrics, $AP_{E}$ and $AP_H$, which take both position and orientation of the gripper into consideration. They can be efficiently evaluated without setting up a robotic system.
    
    The AP metric for 6-DoF grasping evaluates a list of detected grasp poses with different difficulty. We consider a pose is matched with a ground truth pose if both: (1) Translation error to the ground truth is less than 2cm. (2) For hard difficulty ($AP_H$) : Rotation error to the ground truth is less than 5\degree. For easy difficulty ($AP_E$) :  Rotation error to the ground truth is less than 10\degree.
    
    
    We use the rotation metric $d(R_1, R_2) = \arcsin\big(\frac{1}{2\sqrt{2}} || I - R_1 R_2^T ||_F\big)$ in $\mathbb{SO}^3$, which is equivalent to $\phi_3$ in \citep{huynh2009metrics}. The range of values of $d$ is $[0, \frac{\pi}{2}]$ in radians. 
    %
    Since there are infinitely many poses in $\mathbb{SE}^3$ (translation and rotation), we only compute the AP of top-10 grasps returned by the detector with non-maximum suppression.

    \subsection{Experimental Settings}\label{sec:Training data for grasping}
    \textbf{Training Data}.
    We use YCB object set \citep{calli2015ycb} for training, validation, and grasp simulation. We select totally 38 objects that can be grasped by our TOYO CHS2-S68 parallel jaw gripper with the maximum opening width of 98.6mm and sample the ground truth poses by GPG proposed by Pas et al.\citep{ten2017grasp}. Finally, we generated 38843 good grasps, 39154 bad grasps in total, and the training and validation sets are split by the identity of the objects with a ratio of 4:1.
    
    
    \textbf{Baselines}.
    We compare \ours{} to two state-of-the-art, open-sourced baselines, GPD \citep{ten2017grasp} and PointNetGPD \citep{liang2019pointnetgpd}. We train GPD (12-channel) and PointNetGPD with their default setting for 200 epochs. Both PointNetGPD and GPD has 0.94 classification accuracy in validation set.
    
    \textbf{Grasping Tasks}.
    We use PyBullet \citep{coumans2019} to set up the experimental environment which contains the TM5-700 arm, the TOYO CHS2-S68 gripper, a table, the objects to grasp, and one depth camera. We produce the point cloud from the depth camera as the input. We perform two kinds of experiments: single object grasping and clutter removal.
     For single object grasping, an object in the validation set is randomly selected and placed on a table, and then the robot tries to grasp the object. \footnote{All success rates are evaluated from 3000 attempts of grasping to eliminate sampling error. For \emph{clutter removal}, the model needs to grasp all 7 different dropped objects and clear the table. This task is successful only if the table is cleared, and the task is failed if 3 failures happen on the same object.}
    %

    \subsection{Quantitative results}
    \label{sec:Quantitative results}
    \textbf{Main Results}. Table \ref{tab:success_rate} demonstrates the results of single object grasping task in terms of success rate. Our \ours{} significantly outperform all baselines in all object classes (9\% more accuracy overall). Table \ref{tab:clutter_removal_table} shows the complete rate of the clutter removing task, our model (85.33\%) reaches almost twice of accuracy comparing to state-of-the-art method (46.30\%). 
    
    
    
    \textbf{Speed}. One of the key advantage of our end-to-end method is being free from sampling and evaluating multiple grasp candidates. Table \ref{tab:speed} compares the computation time between our method and previous state-of-the-art methods. Our end-to-end method are 20 times faster and preserves impressive accuracy. 
    
    \textbf{Generalization}. Table \ref{tab:AP_num_views} compares the average precision with different numbers of views to test the generalization ability of each method with different numbers of depth cameras. We train all of the models on the 5-view training set and validate on 1-view, 5-view, and 20-view validation set. 
    Both GPD and PointNetGPD suffer from occlusion and get poor performance in 1-view dataset. Our \ours{} surpasses all baselines in a large margin.
    
    \textbf{Robustness}. We analyze the robustness of all methods in sparse point cloud. We limit the maximum number of points for each object. And then test each method with different numbers of points. Table \ref{tab:AP_num_points} shows our \ours{} not only has the best performance but also has the least performance drop given the extremely sparse point cloud.

    \subsection{Ablation study}
    \label{sec:Ablation study} 
    Table \ref{tab:ablation} exhibits the ablation study of our proposed components including the feature extraction layer, coarse-to-fine representation (\emph{C2F}), and the two-fold rotational symmetric representation of roll rotation (\emph{cos-roll}).
    We observe that the naive approach without \emph{C2F} and \emph{cos-roll} seriously suffer from pose ambiguity and led to poor performance. The models adopt only C2F have significantly better performance compared to the naive approach, but still suffer the ambiguity from the two-fold symmetric property of the gripper. The models with both \emph{C2F} and \emph{cos-roll} reduce the pose ambiguity and have the best performance. We also compare the robustness of feature extraction modules in sparse point cloud. The results are shown in supplementary materials.
    
    \begin{table}\label{tab:sog}
    \begin{adjustbox}{max width=\textwidth,center}
    \begin{tabular}{l||c|ccccccc}
    \Xhline{2\arrayrulewidth} 
    Method & 
    \small{Avg.} & 
    \small{\begin{tabular}{@{}c@{}} tomato \\ soup can\end{tabular}} & 
    \small{\begin{tabular}{@{}c@{}} pudding \\ box \end{tabular}} & 
    \small{\begin{tabular}{@{}c@{}} potted \\ meat can \end{tabular}} & 
    \small{orange} & 
    \small{plum} & 
    \small{scissors} & 
    \small{e-cups} \\ 
    \hline PointNetGPD \citep{liang2019pointnetgpd} & 86.03\% & 75.00\% & 94.05\% & 68.59\% & 80.14\% & 94.28\% & 98.44\% & 90.84\% \\ 
    \hline GPD \citep{ten2017grasp} & 87.50\% & 82.60\% & 95.86\% & 76.79\% & 78.83\% & 93.55\% & 98.66\% & 86.83\% \\ 
    \hline S4G$\dagger$ \citep{qin2020s4g} & 69.77\% & \textbf{91.14\%} & 94.32\% & 85.36\% & 56.98\% & 0.00\%\textbf{*} & 75.42\% & 82.21\% \\
    \hline GDN (ours) & \textbf{95.60\%} & 86.83\% & \textbf{98.83\%} & \textbf{95.43\%} & \textbf{100.00\%} & \textbf{99.77\%} & \textbf{100.00\%} & \textbf{95.87\%} \\ 
    \hline 
    \Xhline{2\arrayrulewidth} 
    \end{tabular}
    \end{adjustbox}
    \caption{\label{tab:success_rate} Results of single object grasping experiment. Our \ours{} significantly outperform all state-of-the-art baselines for all objects. The results show that S4G \citep{qin2020s4g} achieves good performance on some objects, such as tomato soup can. However, we observe that S4G$\dagger$ \citep{qin2020s4g} fails to detect grasps on small objects. (*) $\dagger$: We implement the approach according to the original paper. (Section \ref{sec:Quantitative results})}
    \end{table}
    
    \begin{table}
    \centering
    \parbox{0.48\textwidth}{
    \centering
    \resizebox{0.48\columnwidth}{!}{
    \begin{tabular}{c||c|c|c|c|c|c}
    \Xhline{2\arrayrulewidth} &  \multicolumn{2}{c|} {1-view} & \multicolumn{2}{c} {5-view} & \multicolumn{2}{|c} {20-view} \\
    \cline { 2 - 7 }
    & $AP_H$  & $AP_E$ & $AP_H$  & $AP_E$ & $AP_H$  & $AP_E$ \\
    \hline
    PointNetGPD \citep{liang2019pointnetgpd} & 0.1608 & 0.2415 & 0.4514 & 0.5543 & 0.5107 & 0.6070 \\
    GPD \citep{ten2017grasp} & 0.1437 & 0.2295 & 0.3966 & 0.4970 & 0.6032 & 0.7304 \\
    S4G$\dagger$ \citep{qin2020s4g} & 0.4057 & 0.5886 & 0.5186 & 0.7209 & 0.3946 & 0.6265 \\
    GDN (ours) & \textbf{0.6988} & \textbf{0.8622} & \textbf{0.8380} & \textbf{0.9629} & \textbf{0.8200} & \textbf{0.9561} \\
    \Xhline{2\arrayrulewidth} 
    \end{tabular}}
    \caption{\label{tab:AP_num_views}$AP_E$ and $AP_H$ denote the AP for easy and hard difficulties. The result shows that adding more number of views helps improve performance for all methods except S4G$\dagger$ \citep{qin2020s4g}. Our work achieves the highest AP in all cases.}
    }
    \hfill
    \parbox{0.48\textwidth}{
    \centering
    \resizebox{0.48\columnwidth}{!}{
    \begin{tabular}{c||llll}
    \Xhline{2\arrayrulewidth} & \multicolumn{4}{c} {Number of points}  \\
    \cline { 2-5 } 
    & \multicolumn{1}{c}{2048} & \multicolumn{1}{c}{1024} & \multicolumn{1}{c}{512} & \multicolumn{1}{c}{256} \\
    \hline 
    PointNetGPD \citep{liang2019pointnetgpd} & 0.4533 (-0.0\%) & 0.4274 (-5.7\%) & 0.3977 (-12.3\%) & 0.3147 (-30.6\%) \\
    GPD \citep{ten2017grasp} & 0.3838 (-0.0\%) & 0.3019 (-21.3\%) & 0.2006 (-47.7\%) & 0.1589 (-58.6\%) \\
    S4G$\dagger$ \citep{qin2020s4g} & 0.5186 (-0.0\%) & 0.5033 (-3.0\%) & 0.1275 (-75.4\%) & 0.1482 (-71.4\%) \\
    GDN (ours) & 0.8380 (-0.0\%) & 0.8058 (-3.8\%) & 0.7905 (-5.7\%) & 0.7743 (-7.6\%) \\
    \Xhline{2\arrayrulewidth} 
    \end{tabular}}
    \caption{\label{tab:AP_num_points} Each method is evaluated with $AP_H$. The numbers in parentheses are the relative performance drop in percentage. The result shows that our work is robust to sparse point clouds. In contrast, the previous works have significant performance drop in sparse point clouds. }
    }
    \end{table}
    
    \begin{table}
    \centering
    \parbox{0.24\textwidth}{
    \centering
    \resizebox{0.24\columnwidth}{!}{
    \begin{tabular}{l|c}
    \Xhline{2\arrayrulewidth} 
    Method & 
    Complete \\ 
    \hline GDN (ours) & 85.33\% \\ 
    \hline PointNetGPD \citep{liang2019pointnetgpd} & 43.14\% \\ 
    \hline GPD \citep{ten2017grasp} & 46.30\% \\ 
    \hline 
    \Xhline{2\arrayrulewidth} 
    \end{tabular}}
    \caption{\label{tab:clutter_removal_table} Clutter removal experiment.}
    }
    \hfill
    \parbox{0.43\textwidth}{
    \centering
    \resizebox{0.43\columnwidth}{!}{
    \begin{tabular}{l|cccc}
    \Xhline{2\arrayrulewidth} 
    Method & Grasp Proposal & Inference & NMS & Total \\
    \hline GDN (ours) & 0.00 & 0.03 & 0.07 & 0.10 \\
    \hline PointNetGPD \citep{liang2019pointnetgpd} & 0.03 & 2.90 & 0.00 & 2.94 \\
    \hline GPD \citep{ten2017grasp} & 0.03 & 3.29 & 0.00 & 3.32 \\
    \hline 
    \Xhline{2\arrayrulewidth} 
    \end{tabular}}
    \caption{\label{tab:speed} Computation time in second. Our end-to-end approach is over 20 times faster than previous works.}
    }
    \hfill
    \parbox{0.3\textwidth}{
    \centering
    \resizebox{0.23\columnwidth}{!}{
    \begin{tabular}{cc|c}
        \Xhline{2\arrayrulewidth} 
        C2F & cos-roll & $AP_H$ \\
        \hline
        & & 0.0365 \\
        & \ding{51} & 0.0991 \\
        \ding{51} & & 0.1937 \\
        \ding{51} & \ding{51} & 0.8380 \\
        \Xhline{2\arrayrulewidth} 
    \end{tabular}}
    \caption{\label{tab:ablation}The ablation study of proposed components.}
    }
    \end{table}\label{tab:cr}

    \begin{figure}[!hbtp]
    \centering
    \includegraphics[width=\textwidth]{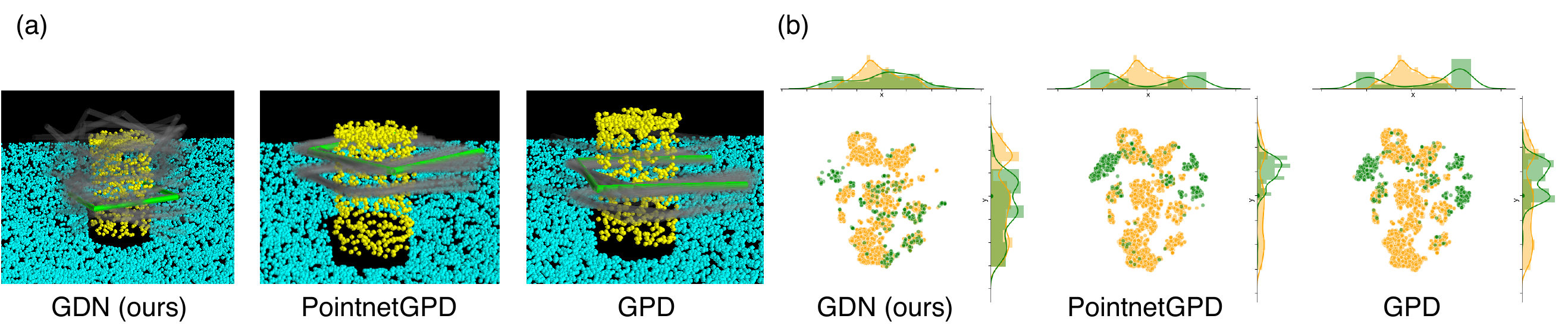}
    \caption{ (a) Grasping in simulated scenes. Gray: The collision-free grasps detected by each method, Green: The executed grasps. (b) The t-SNE visualization to investigate the diversity of grasp poses and the similarity between ground truths and predictions. Yellow dot: ground truths, Green dot: predictions. Both GPD and PointNetGPD have clusters which are far from ground truths. In contrast, ours roughly cover the whole ground truths with more diversity. (Best view in color)}
    \label{fig:tsne}
    \end{figure}
    
    \subsection{Visualization}
    \label{sec:Visualization}
    We observed that both GPD and PointNetGPD suffers from some particular poses and results in failure in 
    the clutter removal tasks. Figure \ref{fig:tsne} reveals that \emph{lack of diversity of grasp poses} might be a reason. From grasping in a simulated scene (left) and the t-SNE visualization (right), the predictions of GPD and PointNetGPD tend to cluster together and cannot cover the whole ground truth, and some of the clusters have no neighboring ground truths. This may explain why GPD and PointNetGPD tend to fail on particular grasp poses and cannot fulfill the task on some objects.
    
    
    \section{Conclusion}
        We investigate the problem of 6-DoF grasp detection among multiple types of objects. The ground-truth grasp distribution inspires us to introduce a coarse-to-fine representation into an end-to-end framework. Our method is able to detect a variety of grasps that have a high coverage on ground-truth grasps. To the best of our knowledge, the C2F representation has not been introduced before in the field related to robotic grasping. Furthermore, the experimental results show that our framework outperforms state-of-the-art two-stage models and can prove that our coarse-to-fine representation is effective in the determination of 6-DoF grasp.

        
    \label{sec:conclusion}

\clearpage

\bibliography{example}  

\acknowledgments{This work was supported in part by the Ministry of Science and Technology, Taiwan, under Grant MOST 109-2634-F-002-032, FIH Mobile Limited, and Qualcomm Technologies, Inc. We benefit from NVIDIA DGX-1 AI Supercomputer and are grateful to the National Center for High-performance Computing. We thank the executives from Techman Robot Inc. for the early discussions.}

\clearpage

\input{Supplementary}

\end{document}

%% file: the_new_approach_section.tex
\section{Problem Formulation}
\label{sec:Problem Formulation}

Given the point cloud of an object in the world coordinate system, we want to find a set of viable 6-DoF grasp poses to pick up the object with a parallel-jaw gripper. We denote $g = (x, y, z, r_x, r_y, r_z) \in \mathbb{R}^6$ as the grasp configuration, where $(x, y, z)$ is the position of a gripper in the world coordinate system, and $(r_x, r_y, r_z)$ is the rotation along x-axis, y-axis, and z-axis. Our goal is to train a robust end-to-end grasp detector to predict possible grasp configurations.    

\section{End-to-end Learning 6-DoF Grasp Detection}
\label{sec:End-to-end Learning 6-DoF Grasp Detection}

    In this section, we present our proposed end-to-end Grasp Detection Network (GDN) with the coarse-to-fine (C2F) grasp representation for detecting a diverse set of 6-DoF grasp poses given point clouds. The overall architecture of our method is illustrated in
    Figure~\ref{fig:2}. 
    
\subsection{Grasp point and relative grasp configuration}
The backbone of our network will subsample the input point cloud into several points, namely \textit{grasp points}, for possible grasp configurations. Then, for each grasp point $p = (x_p, y_p, z_p)$, the final model output will be some relative grasp configurations $g_r$ to $p$ that would be translated into final grasps configurations $g = (x, y, z, r_x, r_y, r_z) \in \mathbb{R}^6$ according to the 3D world coordinate. 

Note that some grasps may be unreachable and rejected by the robot system, so it would be better to have multiple and diverse grasps as outputs. On the other hand, there are usually several ground truth grasp configurations for each grasp point $p$. For this one-to-many problem, we cannot treat these ground truths as separate data, or the model will only learn the average of them. Therefore, we proposed a novel coarse-to-fine (C2F) grasp represnetation bewteen a grasp point and output grasp configurations. 

\subsection{Coarse-to-fine representation (C2F)}


The C2F representation has two concepts: (1) a coarse grasp pose confidence grid and (2) a coarse-to-fine volume. Together it represents some possible grasp configurations around a grasp point and their confidence scores. 

\textbf{Coarse grasp pose confidence grid}.
Instead of output only one grasp configuration per grasp point, our method predicts a confidence grid of multiple grasp configurations. We quantized the grasp orientation along y-axis ($-\frac{\pi}{2} \leq r_y \leq \frac{\pi}{2}$) and z-axis ($-\pi \leq r_z \leq \pi$) of Euler angle into $Ny$ and $Nz$ intervals respectively. For each intervals $i \in [0, N_y-1]$, $j \in [0, N_z-1]$ and point $p$, our models predict a confidence score $c^p_{i,j}$ where $c^p \in \mathbb{R}^{N_y \times N_z}$. This value represents the confidence of finding a good grasp around a coarse orientation $r_y = \frac{\pi}{N_{y}} i - \frac{\pi}{2}$ and $r_z = \frac{2\pi}{N_{z}} j - \pi$. Note that the ground truth orientation along the x-axis (roll) has very low variance, so the x-axis is not included in the confidence grid.

\textbf{Coarse-to-fine volume}.
For each coarse y-axis and z-axis rotation index $i, j$ on the confidence grid, our model also estimates the translation and rotation refinement values. 
The translation, denoted as $(\Delta x^p, \Delta y^p, \Delta z^p)$, is represented by the relative translation from point $p$ to the center of the gripper. For the rotation refinement, our model predicts residual rotation along $y$ and $z$ axis $\Delta {r_y}^p_{i,j}, \Delta {r_z}^p_{i,j}$ based on the coarse rotation determine by $i, j$. 

We directly regress the rotation along $x$-axis. The parallel-jaw gripper is two-fold rotational symmetric along its x-axis in gripper frame. Hence, we parameterize and predict the rotation along the x-axis by cosine and sine function with $2r_x$ as input, avoiding the symmetric ambiguity during regression caused by symmetry. We denote the parameterized rotation by $\theta_{cos} = \cos(2r_x), \theta_{sin} = \sin(2r_x)$.

Combining the confidence grid, the output of the backbone network for a grasp point $p$ is a 3D volume $V^p \in \mathbb{R}^{N_y \times N_z \times 8}$, where $V^p_{i, j}$ is a 8-dimensional vector $(c_{ij}^p, \Delta x^p_{i, j}, \Delta y^p_{i, j}, \Delta z^p_{i, j}, \Delta {r_y}^p_{i,j}, \Delta {r_z}^p_{i,j}, {\theta_{cos}}_{i,j}^p, {\theta_{sin}}_{i,j}^p )$. Each entry $i, j$ of $V^p$ can represent a relative grasp configuration of point $p$ as following: \noindent
\begin{equation}
\begin{aligned}
&\hat{x_p} = \Delta x^p_{i, j}, \quad
\hat{y_p} = \Delta y^p_{i, j}, \quad
\hat{z_p} = \Delta z^p_{i, j}, \quad
\hat{r_x}^p = \arctan(\frac{1-\hat{\theta_{\cos}}^p_{i,j}} {\hat{\theta_{\sin}}^p_{i,j}}), \\
&\hat{r_y}^p = \frac{\pi}{N_{y}} (i + \Delta {r_y}^p_{i,j}) - \frac{\pi}{2}, \quad
\hat{r_z}^p = \frac{2\pi}{N_{z}} (j + \Delta {r_z}^p_{i,j}) - \pi \\
\end{aligned}
\end{equation}

\subsection{Network}
As illustrated in Figure~\ref{fig:2}~(a), given input point cloud, our GDN outputs sub-sampled $N$ grasp points along with the features with shape $(N, Ny \times Nz \times 8$).
We adopt the architecture of PointNet++ \cite{qi2017pointnet++} as backbone network. To enhance the capability of local feature propagation, we replace PointNet layer with with EdgeConv \cite{wang2019dynamic}. We use uniform sampling  for the sampling layer in PointNet++.
The output feature of the backbone network goes through a shared MLP layer, resulting size $(N, Ny \times Nz \times 8)$.

\subsection{Loss Design}
\textbf{Classification Loss}.
We use focal loss \cite{lin2017focal} as grasp classification loss. \noindent
\begin{equation}
\begin{split} 
\mathcal{L}_{cls} &= \frac{1}{|S|} \sum_{\substack{p,i,j}} -\alpha(1-q^p_{i,j})^\gamma \log(q^p_{i,j}) , \\
\end{split}
\end{equation}
where $q^p_{i,j} = c^p_{i,j}$ if $(p,i,j) \in S$ and $q^p_{i,j} = 1-c^p_{i,j}$ otherwise. $S$ is the set of indices whose corresponding poses are close to any ground truth pose. We discuss how a predicted pose will be seen as close to a ground truth in subsection \ref{sec:approach_details}.

\textbf{Rotation Loss}.
The widely used $l1$ or $l2$ distance of Euler angles may suffer from discontinuity and ambiguity. Therefore, we adopt the rotation metric suggested by \citep{huynh2009metrics, wang2019deep}, and calculate the rotation loss in $\mathbb{SO}^3$ to avoid the issues in Euler angle. First, we convert the Euler angle $({r_x}^p_{i,j}, {r_y}^p_{i,j}, {r_z}^p_{i,j})$ to the rotation matrix $R^p_{i,j} \in \mathbb{SO}^3$, and $\widehat{R^p_{i,j}}$ is the predicted rotation of our network, then the rotation loss can be formulated as \noindent
\begin{equation} \begin{split}
\mathcal{L}_{rot} &= \frac{1}{|S|} \sum_{\substack{p, i, j \in S}} {|| I - \widehat{R^p_{i,j}} {R^p_{i,j}}^T ||_F}, \\ &\text{where} \quad p \in P, i \in [0, N_y-1], j \in [0, N_z-1]
\end{split} \end{equation}

\textbf{Translation Loss}. We treat each grasp point $p$ as a point enclosed by some ground truth gripper so that we can regress the translation between the gripper frame and point $p$. And then normalize them by the gripper dimensions to range $[0, 1]$. We formulate a weighted translation loss to minimize translation error, where $\lambda$ is the weighting coefficient and $x,y,z$ are the normalized translation as mentioned above: \noindent 
\begin{equation} \begin{split} 
\mathcal{L}_{trans} = \frac{1}{|S|} \sum_{p,i,j \in S} {  \lambda_x ||x^p_{i,j} - \hat{x}^p_{i,j}||_1 + \lambda_y ||y^p_{i,j} - \hat{y}^p_{i,j}||_1 + \lambda_z ||z^p_{i,j} - \hat{z}^p_{i,j}||_1 } 
\end{split} \end{equation}

Finally, we can combine each term and formulate the total loss with weighting coefficients as \noindent 
\begin{equation} \begin{split} 
\mathcal{L}_{total} = \lambda_{cls} \mathcal{L}_{cls} + \lambda_{rot} \mathcal{L}_{rot} + \mathcal{L}_{trans}
\end{split} \end{equation}
    
\subsection{Other Details} \label{sec:approach_details}
When calculating the classification loss, a predicted pose on $(p, i, j)$ is said to be close to a ground truth pose if and only if its grasp point $p$ is enclosed by the gripper using that ground truth pose, and its coarse orientation $i, j$ is closest to that ground truth orientation among all $N_y \times N_z$ coarse poses in the grasp point $p$.
When calculating the rotation and translation losses, if multiple ground truth grasp configurations are corresponding to the same predicted coarse pose $p, i, j$, we keep the one with the smallest residual for calculating the refinement losses.

In the inference stage, we rank the predicted poses by their confidence score $c^p_{i, j}$ for all $p, i, j$ and apply the non-maximum suppression (NMS) algorithm as used in \cite{ren2015faster} to filter out poses with high overlap. In the NMS algorithm, two poses is considered to overlap with each other if their difference of traslation and rotation are less than 2cm and 5\degree, respectively.

%% file: Supplementary.tex

\section*{Appendix A \ \  Comparison to S4G \citep{qin2020s4g}}

\begin{figure}[!hbtp]
    \centering
    \includegraphics[width=0.9\textwidth]{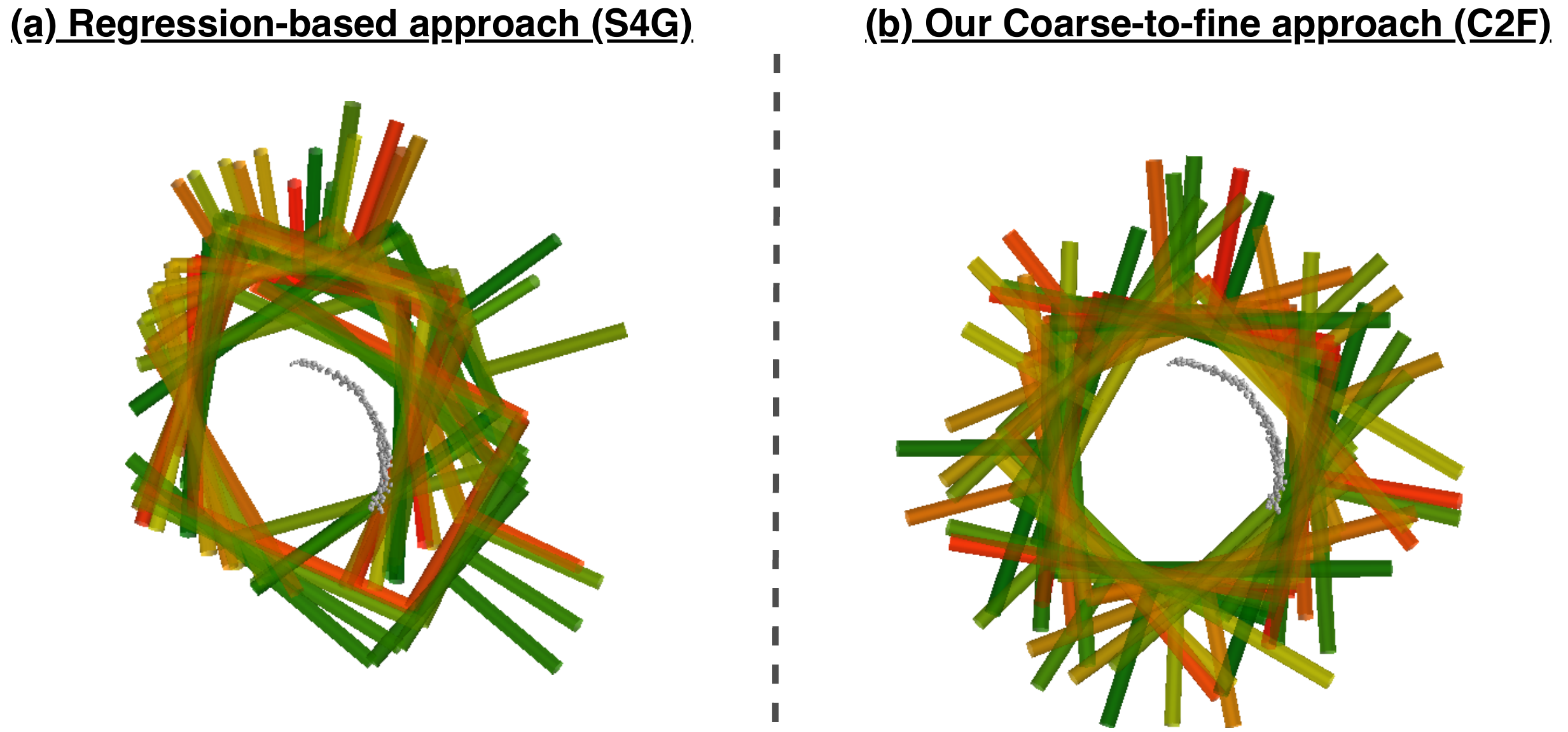}
    \caption{ Visualization of grasp detection. The color of the grippers represents the confidence scores of grasp poses. Green is the highest and red is the lowest. (a) The regression-based approach (S4G) leads to some loss of pose diversity. (b) Our coarse-to-fine approach can keep the diversity of orientation, which is crucial for collision avoidance with adjacent objects based on partial point clouds.}%
    \label{fig:visualize_gdn_vs_s4g}%
\end{figure}

\begin{figure}[!hbtp]
    \centering
    \begin{tabular}{rm{0.22\textwidth}m{0.22\textwidth}}
      & \multicolumn{1}{c}{GDN} & \multicolumn{1}{c}{S4G \citep{qin2020s4g}} \\
      \begin{tabular}{@{}c@{}}potted \\ meat can\end{tabular} &
      \includegraphics[width=0.22\textwidth]{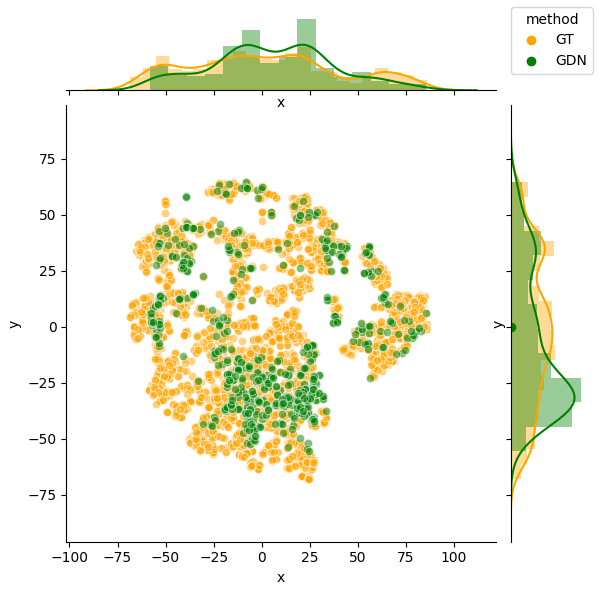} &
      \includegraphics[width=0.22\textwidth]{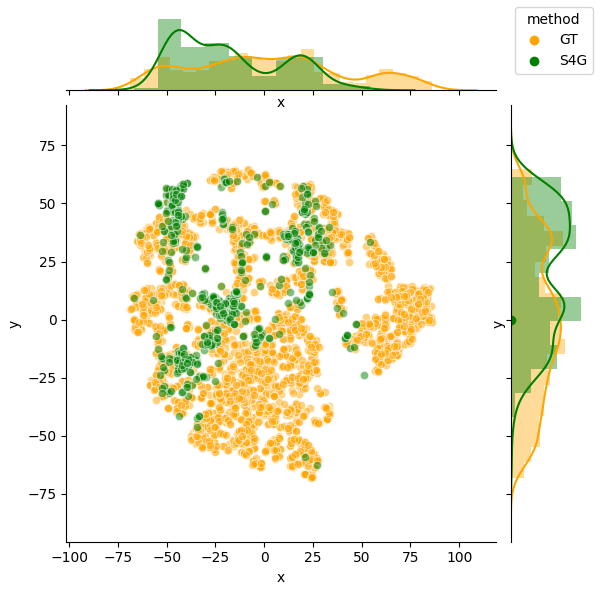} \\
      cup  &
      \includegraphics[width=0.22\textwidth]{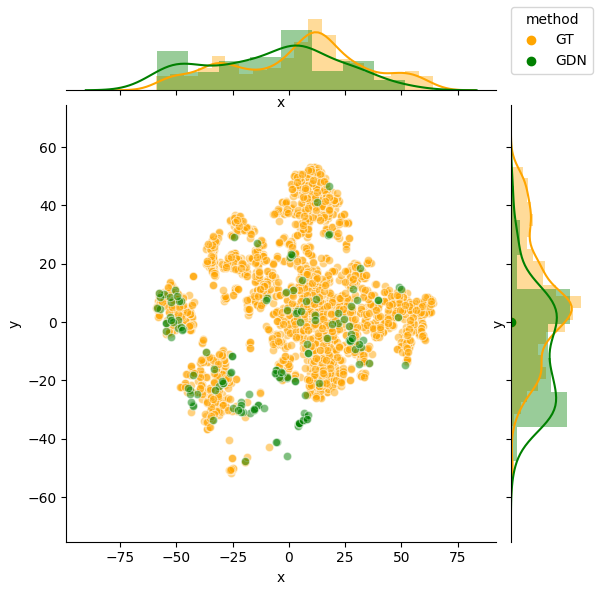} &
      \includegraphics[width=0.22\textwidth]{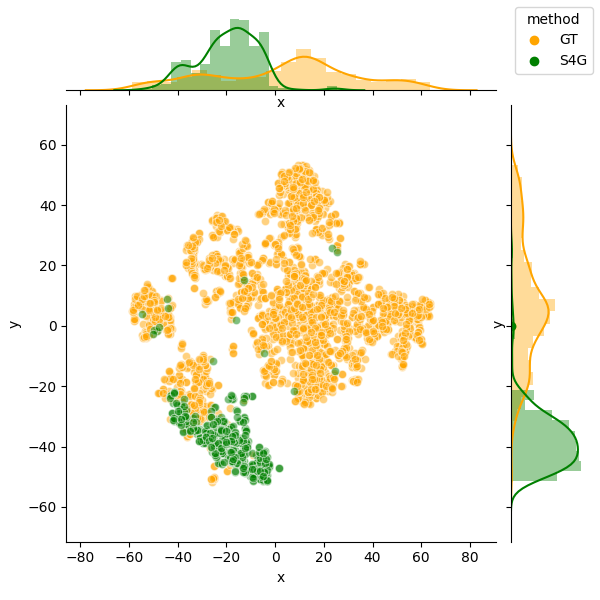} \\
    \end{tabular}
    \caption{The t-SNE visualization to investigate the diversity ofgrasp poses and the similarity between ground truths and predictions.  Yellow dot: ground truths, Green dot: predictions. Ours roughly cover the ground truths and has higher diversity compared to S4G. (Best view in color)}
    \label{fig:tsne_s4g}
\end{figure}

To further compare to other end-to-end approaches that are recently proposed, we evaluate the performance of the S4G model \citep{qin2020s4g} on our dataset. We show that our C2F representation is effective in detection of 6-DoF grasp in Table~\ref{tab:success_rate}-\ref{tab:AP_num_points} and explain why C2F representation is better than the representation proposed in S4G via Figure \ref{fig:visualize_gdn_vs_s4g} and  \ref{fig:tsne_s4g}. 

\clearpage
\section*{Appendix B \ \  Ablation Study}

We investigate the effectiveness of different feature extractors in our end-to-end model in Table \ref{tab:AP_num_points_ablation}. The results show that the model with EdgeConv module performs better than the model with  PointNet module for 6-DoF grasp detection.

\begin{table}[!hbtp]
    \centering
    \begin{tabular}{c|c||llll}
    \Xhline{2\arrayrulewidth} \multirow{2}{*}{Method} & \multirow{2}{*}{\begin{tabular}{@{}c@{}} Feature \\ Extractor \end{tabular}} & \multicolumn{4}{c} {Number of points}  \\
    \cline { 3-6 } 
    & & \multicolumn{1}{c}{2048} & \multicolumn{1}{c}{1024} & \multicolumn{1}{c}{512} & \multicolumn{1}{c}{256} \\
    \hline 
    GDN (ours) & PointNet & 0.8362 (-0.0\%) & 0.8021 (-4.1\%) & 0.4593 (-45.1\%) & 0.4196 (-49.8\%) \\
    GDN (ours) & EdgeConv & 0.8380 (-0.0\%) & 0.8058 (-3.8\%) & 0.7905 (-5.7\%) & 0.7743 (-7.6\%) \\
    \Xhline{2\arrayrulewidth} 
    \end{tabular}
    \caption{\label{tab:AP_num_points_ablation} Each method is evaluated with $AP_H$. The numbers in parentheses are the relative performance drop in percentage. The method which use PointNet as feature extractor has a huge performance drop in sparse point clouds. }
\end{table}

\section*{Appendix C \ \  Clutter Removal Process}
\begin{figure}[!hbtp]
    \begin{adjustbox}{max width=\textwidth,center}
    \begin{tabular}{cccc}
      \includegraphics{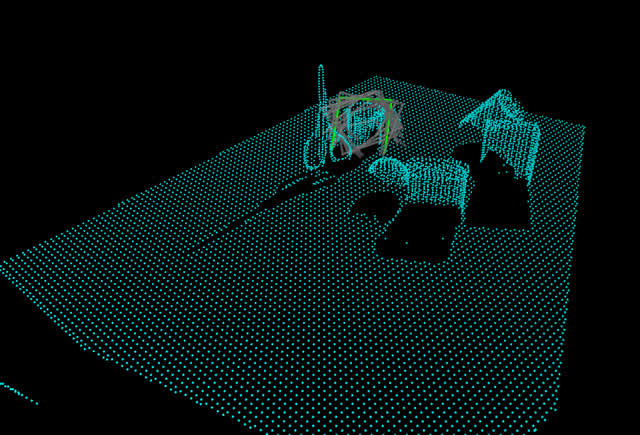} &
      \includegraphics{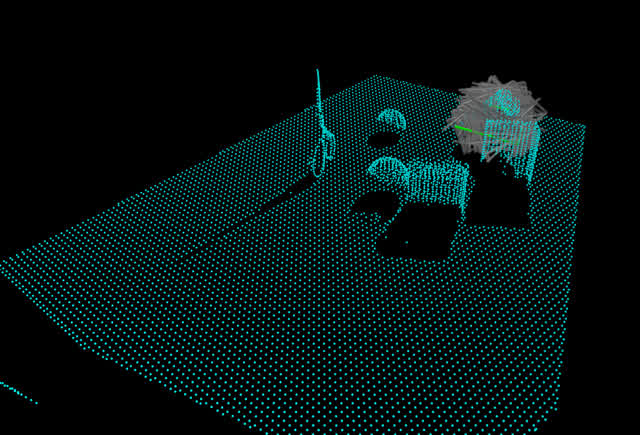} &
      \includegraphics{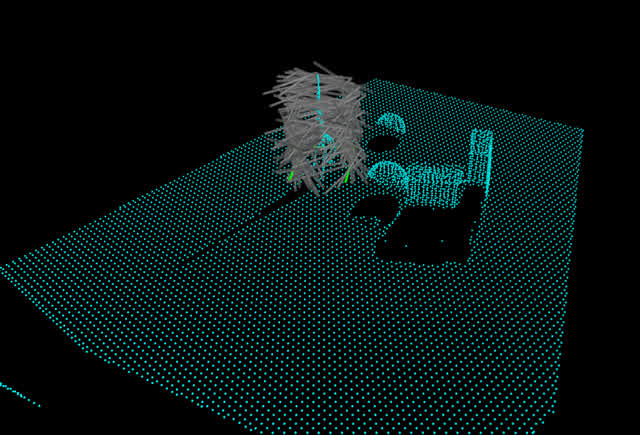} &
      \includegraphics{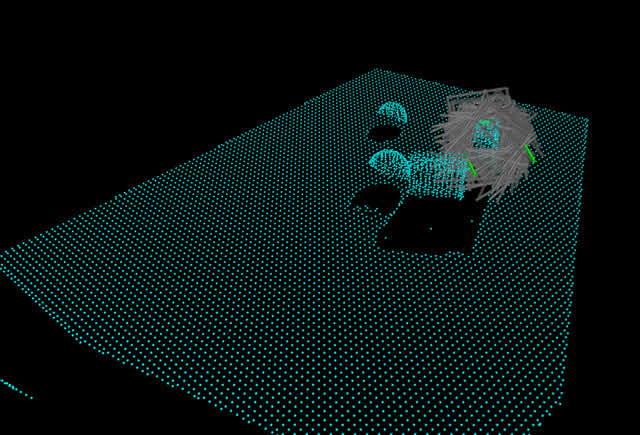} \\ \\
      \includegraphics{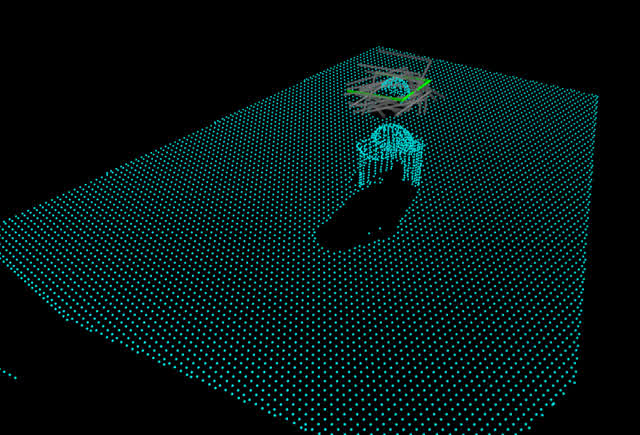} &
      \includegraphics{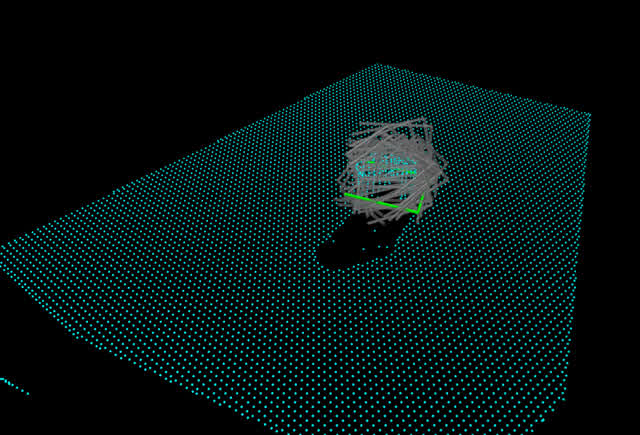} &
      \includegraphics{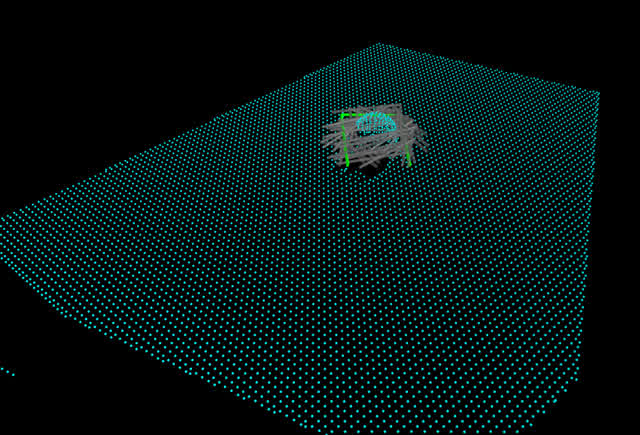}
    \end{tabular}
    \end{adjustbox}
    \caption{Step-by-step visualization of clutter removal in simulated scene. The collision-free grasps chosen by our GDN are annotated in gray, and the executed grasps are annotate in green. (Zoom-in to observe the details.)}
    \label{fig:simulated_scene}
\end{figure}

We show our clutter object removal process with a step-by-step visualization in Figure \ref{fig:simulated_scene}.

\section*{Appendix D \ \  Network Architecture Details}

The detailed network architecture of our \ours{} is shown below:

EdgeConvs([32,64], 20) $\rightarrow$ Subsample(1024) $\rightarrow$ EdgeConvs([128,128,256], 10) $\rightarrow$
Subsample(256) $\rightarrow$
EdgeConvs([512,512,1024], 10) $\rightarrow$
Subsample(64) $\rightarrow$
Conv1D(4800) $\rightarrow$ Reshape(64,24,25,8)
, where EdgeConvs([$c_1,c_2$,...], $K$) is a series of edge convolution layers with $c_i, i \in 1, 2, ..., n$ output channels and $K$ nearest neighbors, Subsample($N$) is a sub-sample layer which uniformly samples the input data to $N$ points. We use 1D convolution with 1x1 kernel to implement shared MLP.